\batchmode
\makeatletter
\def\input@path{{/home/fyzhu/DATA/Dropbox/self_Folder/myWorksOnDropboxs/201702_MICCAI_GroupDrivenRL_4_mHealth_batchLearning//}}
\makeatother
\documentclass{llncs}
\usepackage[latin9]{inputenc}
\usepackage{color}
\definecolor{page_backgroundcolor}{rgb}{0.890625, 0.929688, 0.804688}
\pagecolor{page_backgroundcolor}
\usepackage{array}
\usepackage{bm}
\usepackage{multirow}
\usepackage{amsmath}
\usepackage{amssymb}
\usepackage{graphicx}
\usepackage[unicode=true,pdfusetitle,
 bookmarks=true,bookmarksnumbered=false,bookmarksopen=false,
 breaklinks=true,pdfborder={0 0 0},backref=false,colorlinks=true]
 {hyperref}

\makeatletter

\providecommand{\tabularnewline}{\\}

\usepackage{llncsdoc}
\usepackage{makeidx}\usepackage{epsfig}
\usepackage{url}

\usepackage{epsfig}
\usepackage{algorithm}
\usepackage{algorithmic}
\usepackage{subfigure}
\usepackage{caption}
\usepackage{epsfig}
\usepackage{epstopdf}

\makeatother

\begin{document}
\global\long\def\mtbfA{\mathbf{A}}
 \global\long\def\mtbfa{\mathbf{a}}
 \global\long\def\mebfA{\bar{\mtbfA}}
 \global\long\def\mebfa{\bar{\mtbfa}}

\global\long\def\mhbfA{\widehat{\mathbf{A}}}
 \global\long\def\mhbfa{\widehat{\mathbf{a}}}
 \global\long\def\mtcalA{\mathcal{A}}
 \global\long\def\mtbbA{\mathbb{A}}

\global\long\def\mtbfB{\mathbf{B}}
 \global\long\def\mtbfb{\mathbf{b}}
 \global\long\def\mebfB{\bar{\mtbfB}}
 \global\long\def\mebfb{\bar{\mtbfb}}

\global\long\def\mhbfB{\widehat{\mathbf{B}}}
 \global\long\def\mhbfb{\widehat{\mathbf{b}}}
 \global\long\def\mtcalB{\mathcal{B}}
 \global\long\def\mtbbB{\mathbb{B}}

\global\long\def\mtbfC{\mathbf{C}}
 \global\long\def\mtbfc{\mathbf{c}}
 \global\long\def\mebfC{\bar{\mtbfC}}
 \global\long\def\mebfc{\bar{\mtbfc}}

\global\long\def\mhbfC{\widehat{\mathbf{C}}}
 \global\long\def\mhbfc{\widehat{\mathbf{c}}}
 \global\long\def\mtcalC{\mathcal{C}}
 \global\long\def\mtbbC{\mathbb{C}}

\global\long\def\mtbfD{\mathbf{D}}
 \global\long\def\mtbfd{\mathbf{d}}
 \global\long\def\mebfD{\bar{\mtbfD}}
 \global\long\def\mebfd{\bar{\mtbfd}}

\global\long\def\mhbfD{\widehat{\mathbf{D}}}
 \global\long\def\mhbfd{\widehat{\mathbf{d}}}
 \global\long\def\mtcalD{\mathcal{D}}
 \global\long\def\mtbbD{\mathbb{D}}

\global\long\def\mtbfE{\mathbf{E}}
 \global\long\def\mtbfe{\mathbf{e}}
 \global\long\def\mebfE{\bar{\mtbfE}}
 \global\long\def\mebfe{\bar{\mtbfe}}

\global\long\def\mhbfE{\widehat{\mathbf{E}}}
 \global\long\def\mhbfe{\widehat{\mathbf{e}}}
 \global\long\def\mtcalE{\mathcal{E}}
 \global\long\def\mtbbE{\mathbb{E}}

\global\long\def\mtbfF{\mathbf{F}}
 \global\long\def\mtbff{\mathbf{f}}
 \global\long\def\mebfF{\bar{\mathbf{F}}}
 \global\long\def\mebff{\bar{\mathbf{f}}}

\global\long\def\mhbfF{\widehat{\mathbf{F}}}
 \global\long\def\mhbff{\widehat{\mathbf{f}}}
 \global\long\def\mtcalF{\mathcal{F}}
 \global\long\def\mtbbF{\mathbb{F}}

\global\long\def\mtbfG{\mathbf{G}}
 \global\long\def\mtbfg{\mathbf{g}}
 \global\long\def\mebfG{\bar{\mathbf{G}}}
 \global\long\def\mebfg{\bar{\mathbf{g}}}

\global\long\def\mhbfG{\widehat{\mathbf{G}}}
 \global\long\def\mhbfg{\widehat{\mathbf{g}}}
 \global\long\def\mtcalG{\mathcal{G}}
 \global\long\def\mtbbG{\mathbb{G}}

\global\long\def\mtbfH{\mathbf{H}}
 \global\long\def\mtbfh{\mathbf{h}}
 \global\long\def\mebfH{\bar{\mathbf{H}}}
 \global\long\def\mebfh{\bar{\mathbf{h}}}

\global\long\def\mhbfH{\widehat{\mathbf{H}}}
 \global\long\def\mhbfh{\widehat{\mathbf{h}}}
 \global\long\def\mtcalH{\mathcal{H}}
 \global\long\def\mtbbH{\mathbb{H}}

\global\long\def\mtbfI{\mathbf{I}}
 \global\long\def\mtbfi{\mathbf{i}}
 \global\long\def\mebfI{\bar{\mathbf{I}}}
 \global\long\def\mebfi{\bar{\mathbf{i}}}

\global\long\def\mhbfI{\widehat{\mathbf{I}}}
 \global\long\def\mhbfi{\widehat{\mathbf{i}}}
 \global\long\def\mtcalI{\mathcal{I}}
 \global\long\def\mtbbI{\mathbb{I}}

\global\long\def\mtbfJ{\mathbf{J}}
 \global\long\def\mtbfj{\mathbf{j}}
 \global\long\def\mebfJ{\bar{\mathbf{J}}}
 \global\long\def\mebfj{\bar{\mathbf{j}}}

\global\long\def\mhbfJ{\widehat{\mathbf{J}}}
 \global\long\def\mhbfj{\widehat{\mathbf{j}}}
 \global\long\def\mtcalJ{\mathcal{J}}
 \global\long\def\mtbbJ{\mathbb{J}}

\global\long\def\mtbfK{\mathbf{K}}
 \global\long\def\mtbfk{\mathbf{k}}
 \global\long\def\mebfK{\bar{\mathbf{K}}}
 \global\long\def\mebfk{\bar{\mathbf{k}}}

\global\long\def\mhbfK{\widehat{\mathbf{K}}}
 \global\long\def\mhbfk{\widehat{\mathbf{k}}}
 \global\long\def\mtcalK{\mathcal{K}}
 \global\long\def\mtbbK{\mathbb{K}}

\global\long\def\mtbfL{\mathbf{L}}
 \global\long\def\mtbfl{\mathbf{l}}
 \global\long\def\mebfL{\bar{\mathbf{L}}}
 \global\long\def\mebfl{\bar{\mathbf{l}}}

\global\long\def\mhbfL{\widehat{\mathbf{K}}}
 \global\long\def\mhbfl{\widehat{\mathbf{k}}}
 \global\long\def\mtcalL{\mathcal{L}}
 \global\long\def\mtbbL{\mathbb{L}}

\global\long\def\mtbfM{\mathbf{M}}
 \global\long\def\mtbfm{\mathbf{m}}
 \global\long\def\mebfM{\bar{\mathbf{M}}}
 \global\long\def\mebfm{\bar{\mathbf{m}}}

\global\long\def\mhbfM{\widehat{\mathbf{M}}}
 \global\long\def\mhbfm{\widehat{\mathbf{m}}}
 \global\long\def\mtcalM{\mathcal{M}}
 \global\long\def\mtbbM{\mathbb{M}}

\global\long\def\mtbfN{\mathbf{N}}
 \global\long\def\mtbfn{\mathbf{n}}
 \global\long\def\mebfN{\bar{\mathbf{N}}}
 \global\long\def\mebfn{\bar{\mathbf{n}}}

\global\long\def\mhbfN{\widehat{\mathbf{N}}}
 \global\long\def\mhbfn{\widehat{\mathbf{n}}}
 \global\long\def\mtcalN{\mathcal{N}}
 \global\long\def\mtbbN{\mathbb{N}}

\global\long\def\mtbfO{\mathbf{O}}
 \global\long\def\mtbfo{\mathbf{o}}
 \global\long\def\mebfO{\bar{\mathbf{O}}}
 \global\long\def\mebfo{\bar{\mathbf{o}}}

\global\long\def\mhbfO{\widehat{\mathbf{O}}}
 \global\long\def\mhbfo{\widehat{\mathbf{o}}}
 \global\long\def\mtcalO{\mathcal{O}}
 \global\long\def\mtbbO{\mathbb{O}}

\global\long\def\mtbfP{\mathbf{P}}
 \global\long\def\mtbfp{\mathbf{p}}
 \global\long\def\mebfP{\bar{\mathbf{P}}}
 \global\long\def\mebfp{\bar{\mathbf{p}}}

\global\long\def\mhbfP{\widehat{\mathbf{P}}}
 \global\long\def\mhbfp{\widehat{\mathbf{p}}}
 \global\long\def\mtcalP{\mathcal{P}}
 \global\long\def\mtbbP{\mathbb{P}}

\global\long\def\mtbfQ{\mathbf{Q}}
 \global\long\def\mtbfq{\mathbf{q}}
 \global\long\def\mebfQ{\bar{\mathbf{Q}}}
 \global\long\def\mebfq{\bar{\mathbf{q}}}

\global\long\def\mhbfQ{\widehat{\mathbf{Q}}}
 \global\long\def\mhbfq{\widehat{\mathbf{q}}}
\global\long\def\mtcalQ{\mathcal{Q}}
 \global\long\def\mtbbQ{\mathbb{Q}}

\global\long\def\mtbfR{\mathbf{R}}
 \global\long\def\mtbfr{\mathbf{r}}
 \global\long\def\mebfR{\bar{\mathbf{R}}}
 \global\long\def\mebfr{\bar{\mathbf{r}}}

\global\long\def\mhbfR{\widehat{\mathbf{R}}}
 \global\long\def\mhbfr{\widehat{\mathbf{r}}}
\global\long\def\mtcalR{\mathcal{R}}
 \global\long\def\mtbbR{\mathbb{R}}

\global\long\def\mtbfS{\mathbf{S}}
 \global\long\def\mtbfs{\mathbf{s}}
 \global\long\def\mebfS{\bar{\mathbf{S}}}
 \global\long\def\mebfs{\bar{\mathbf{s}}}

\global\long\def\mhbfS{\widehat{\mathbf{S}}}
 \global\long\def\mhbfs{\widehat{\mathbf{s}}}
\global\long\def\mtcalS{\mathcal{S}}
 \global\long\def\mtbbS{\mathbb{S}}

\global\long\def\mtbfT{\mathbf{T}}
 \global\long\def\mtbft{\mathbf{t}}
 \global\long\def\mebfT{\bar{\mathbf{T}}}
 \global\long\def\mebft{\bar{\mathbf{t}}}

\global\long\def\mhbfT{\widehat{\mathbf{T}}}
 \global\long\def\mhbft{\widehat{\mathbf{t}}}
 \global\long\def\mtcalT{\mathcal{T}}
 \global\long\def\mtbbT{\mathbb{T}}

\global\long\def\mtbfU{\mathbf{U}}
 \global\long\def\mtbfu{\mathbf{u}}
 \global\long\def\mebfU{\bar{\mathbf{U}}}
 \global\long\def\mebfu{\bar{\mathbf{u}}}

\global\long\def\mhbfU{\widehat{\mathbf{U}}}
 \global\long\def\mhbfu{\widehat{\mathbf{u}}}
 \global\long\def\mtcalU{\mathcal{U}}
 \global\long\def\mtbbU{\mathbb{U}}

\global\long\def\mtbfV{\mathbf{V}}
 \global\long\def\mtbfv{\mathbf{v}}
 \global\long\def\mebfV{\bar{\mathbf{V}}}
 \global\long\def\mebfv{\bar{\mathbf{v}}}

\global\long\def\mhbfV{\widehat{\mathbf{V}}}
 \global\long\def\mhbfv{\widehat{\mathbf{v}}}
\global\long\def\mtcalV{\mathcal{V}}
 \global\long\def\mtbbV{\mathbb{V}}

\global\long\def\mtbfW{\mathbf{W}}
 \global\long\def\mtbfw{\mathbf{w}}
 \global\long\def\mebfW{\bar{\mathbf{W}}}
 \global\long\def\mebfw{\bar{\mathbf{w}}}

\global\long\def\mhbfW{\widehat{\mathbf{W}}}
 \global\long\def\mhbfw{\widehat{\mathbf{w}}}
 \global\long\def\mtcalW{\mathcal{W}}
 \global\long\def\mtbbW{\mathbb{W}}

\global\long\def\mtbfX{\mathbf{X}}
 \global\long\def\mtbfx{\mathbf{x}}
 \global\long\def\mebfX{\bar{\mtbfX}}
 \global\long\def\mebfx{\bar{\mtbfx}}

\global\long\def\mhbfX{\widehat{\mathbf{X}}}
 \global\long\def\mhbfx{\widehat{\mathbf{x}}}
 \global\long\def\mtcalX{\mathcal{X}}
 \global\long\def\mtbbX{\mathbb{X}}

\global\long\def\mtbfY{\mathbf{Y}}
 \global\long\def\mtbfy{\mathbf{y}}
\global\long\def\mebfY{\bar{\mathbf{Y}}}
 \global\long\def\mebfy{\bar{\mathbf{y}}}

\global\long\def\mhbfY{\widehat{\mathbf{Y}}}
 \global\long\def\mhbfy{\widehat{\mathbf{y}}}
 \global\long\def\mtcalY{\mathcal{Y}}
 \global\long\def\mtbbY{\mathbb{Y}}

\global\long\def\mtbfZ{\mathbf{Z}}
 \global\long\def\mtbfz{\mathbf{z}}
 \global\long\def\mebfZ{\bar{\mathbf{Z}}}
 \global\long\def\mebfz{\bar{\mathbf{z}}}

\global\long\def\mhbfZ{\widehat{\mathbf{Z}}}
 \global\long\def\mhbfz{\widehat{\mathbf{z}}}
\global\long\def\mtcalZ{\mathcal{Z}}
 \global\long\def\mtbbZ{\mathbb{Z}}

\global\long\def\mtth{\text{th}}

\global\long\def\mtbfzero{\mathbf{0}}
 \global\long\def\mtbfone{\mathbf{1}}

\global\long\def\mttrace{\text{Tr}}

\global\long\def\mttotalVariation{\text{TV}}

\global\long\def\mtexpect{\mathbb{E}}

\global\long\def\mtdet{\text{det}}

\global\long\def\mtvec{\mathbf{\text{vec}}}

\global\long\def\mtvar{\mathbf{\text{var}}}

\global\long\def\mtcov{\mathbf{\text{cov}}}

\global\long\def\mtsubTo{\mathbf{\text{s.t.}}}

\global\long\def\mtfor{\text{for}}

\global\long\def\mtrank{\text{rank}}

\global\long\def\mtrankn{\text{rankn}}

\global\long\def\mtdiag{\mathbf{\text{diag}}}

\global\long\def\mtsign{\mathbf{\text{sign}}}

\global\long\def\mtloss{\mathbf{\text{loss}}}

\global\long\def\mtwhen{\text{when}}

\global\long\def\mtwhere{\text{where}}

\global\long\def\mtif{\text{if}}

\title{Group-driven Reinforcement Learning for Personalized mHealth Intervention}

\author{Feiyun Zhu$^{1,2}$, Jun Guo$^{2}$, Zheng Xu$^{1}$, Peng Liao$^{2}$
and Junzhou Huang$^{1}$}

\institute{$^{1}$ Department of CSE, University of Texas at Arlington, TX,
76013, USA\\
$^{2}$ Department of Statistics, Univeristy of Michigan, Ann Arbor,
MI 48109, USA}
\maketitle
\begin{abstract}
Due to the popularity of smartphones and wearable devices nowadays,
mobile health (mHealth) technologies are promising to bring positive
and wide impacts on people's health. State-of-the-art decision-making
methods for mHealth rely on some ideal assumptions. Those methods
either assume that the users are completely homogenous or completely
heterogeneous. However, in reality, a user might be similar with some,
but not all, users. In this paper, we propose a novel group-driven
reinforcement learning method for the mHealth. We aim to understand
how to share information among similar users to better convert the
limited user information into sharper learned RL policies. Specifically,
we employ the K-means clustering method to group users based on their
trajectory information similarity and learn a shared RL policy for
each group. Extensive experiment results have shown that our method
can achieve clear gains over the state-of-the-art RL methods for mHealth. 
\end{abstract}

\section{Introduction}

In the wake of the vast population of smart devices\footnote{smartphones and wearable devices such as the Fitbit Fuelband and Jawbone
etc. } users worldwide, mobile health (mHealth) technologies become increasingly
popular among the scientist communities. The goal of mHealth is to
use smart devices as great platforms to collect and analyze raw data
(weather, location, social activity, stress, etc.). Based on that,
the aim is to provide in-time interventions to device users according
to their ongoing status and changing needs, helping users to lead
healthier lives, such as reducing the alcohol abuse\ \cite{Gustafson_2014_JAMA_drinking,Witkiewitz_2014_JAB_drinkingSmoking}
and the obesity management\ \cite{Patrick_2009_JMIR_weightManagement}.

Formally, the tailoring of mHealth intervention is modeled as a sequential
decision making (SDM) problem. It aims to learn the optimal decision
rule to decide when, where and how to deliver interventions\ \cite{huitian_2014_NIPS_ActCriticBandit4JITAI,PengLiao_2015_Proposal_offPolicyRL,SusanMurphy_2016_CORR_BatchOffPolicyAvgRwd,fyZhu_2017_arXiv_CohesionDrivenActorCriticRL,fyzhu_2017_RLDM_WarmStart}
to best serve users. This is a new research topic. Currently, there
are two types of reinforcement learning (RL) methods for mHealth with
distinct assumptions: (a) the off-policy, batch RL\ \cite{PengLiao_2015_Proposal_offPolicyRL,SusanMurphy_2016_CORR_BatchOffPolicyAvgRwd}
assumes that all users in the mHealth are completely homogenous: they
share all information and learn an identical RL for all the users;
(b) the on-policy, online RL\ \cite{huitian_2014_NIPS_ActCriticBandit4JITAI,huitian_2016_PhdThesis_actCriticAlgorithm,fyzhu_2017_RLDM_WarmStart}
assumes that all users are completely different: they share no information
and run a separate RL for each user. The above assumptions are good
as a start for the mHealth study. However, when mHealth are applied
to more practical situations, they have the following drawbacks: (a)
the off-policy, batch RL method ignore the fact that the behavior
of all users may be too complicated to be modeled with an identical
RL, which leads to potentially large biases in the learned policy;
(b) for the on-policy, online RL method, an individual user\textquoteright s
trajectory data is hardly enough to support a separate RL learning,
which is likely to result in unstable policies that contain lots of
variances.

A more realistic assumption lies between the above two extremes: a
user may be similar to some, but not all, users and similar users
tend to have similar behaviors\ \cite{fyzhu_2014_IJPRS_SSNMF,haichangLi_2016_IJRS_LablePropagationHyperClassification}.
In this paper, we propose a novel group driven RL for the mHealth.
It is in an actor-critic setting\ \cite{Grondman_2012_IEEEts_surveyOfActorCriticRL}.
The core idea is to find the similarity (cohesion) network for the
users. Specifically, we employ the clustering method to mine the group
information. Taking the group information into consideration, we
learn $K$ (i.e. the number of groups) shared RLs for $K$ groups
of users respectively; each RL learning procedure makes use of all
the data in that group. Such implementation balances the conflicting
goals of reducing the complexity of data while enriching the number
of samples for each RL learning process.

\section{Preliminaries}

The Markov Decision Process (MDP) provides a mathematical tool to
model the dynamic system\ \cite{Geist_2013_TNNLS_RL_valueFunctionApproximation,Grondman_2012_IEEEts_surveyOfActorCriticRL,LichaoWang_2011_MICCAI_RL_4_Segmentation}.
It is defined as a 5-tuple $\left\{ \mtcalS,\mtcalA,P,R,\gamma\right\} $,
where $\mtcalS$ is the state space and $\mtcalA$ is the (finite)
action space. The state transition model $\mtcalP:\mtcalS\times\mtcalA\times\mtcalS\mapsto\left[0,1\right]$
indicates the probability of transiting from one state $s$ to another
$s'$ under a given action $a$. $\mtcalR:\mtcalS\times\mtcalA\mapsto\mtbbR$
is the corresponding reward, which is assumed to be bounded over the
state and action spaces. $\gamma\in[0,1)$ is a discount factor that
reduces the influence of future rewards. The stochastic policy $\pi\left(\cdot\mid s\right)$
determines how the agent acts with the system by providing each state
$s$ with a probability distribution over all the possible actions.
We consider the parameterized stochastic policy, i.e., $\pi_{\theta}\left(a\mid s\right)$,
where $\theta$ is the unkown coefficients.

Formally, the quality of a policy $\pi$ is evaluated by a value function
$Q^{\pi}\left(s,a\right)\in\mtbbR^{\left|\mtcalS\right|\times\left|\mtcalA\right|}$.
It specifies the total amount of rewards (called return) an agent
can achieve when starting from state $s$, first choosing action $a$
and then following the policy $\pi$. It is defined as follows\ \cite{Grondman_2012_IEEEts_surveyOfActorCriticRL}:
\begin{equation}
Q^{\pi}\left(s,a\right)=\mtexpect_{a_{i}\sim\pi,s_{i}\sim\mtcalP}\left\{ \sum_{t=0}^{\infty}\gamma^{t}\mtcalR\left(s_{t},a_{t}\right)\mid s_{0}=s,a_{0}=a\right\} .\label{eq:Q_value}
\end{equation}

The goal of various RL methods is to learn an optimal policy $\pi^{*}$
that maximizes the Q-value for all the state-action pairs\ \cite{Geist_2013_TNNLS_RL_valueFunctionApproximation}.
The objective is $\pi_{\theta^{*}}=\arg\max_{\theta}\widehat{J}$$\left(\theta\right)$.
Such procedure is called the actor updating\ \cite{Grondman_2012_IEEEts_surveyOfActorCriticRL}.
Here 
\begin{equation}
\widehat{J}\left(\theta\right)=\sum_{s\in\mtcalS}d_{\text{ref}}\left(s\right)\sum_{a\in\mtcalA}\pi_{\theta}\left(a\mid s\right)Q^{\pi_{\theta}}\left(s,a\right),\label{eq:obj_actorUpdating}
\end{equation}
where $d_{\text{ref}}\left(s\right)$ is a reference distribution
over states; $Q^{\pi_{\theta}}\left(s,a\right)$ is the value for
the parameterized policy $\pi_{\theta}$. It is obvious that we need
the estimation of $Q^{\pi_{\theta}}\left(s,a\right)$ (i.e. critic
updating) to determine the objective function\ \eqref{eq:obj_actorUpdating}.

Since in mHealth the state space is very large, it is impossible to
directly estimate the Q-value because of the high storage requirement.
Instead, the linear approximation alleviates this problem by assuming
that $Q^{\pi}$ is in a low dimensional space\ \cite{Sutton_2012_Book_ReinforcementLearning}:
$Q_{\mtbfw}=\mtbfw^{T}\mtbfx\left(s,a\right)\approx Q^{\pi}$ where
$\mtbfx\left(s,a\right)$ is a feature processing step that combines
the information in the state and action.

\section{Cohesion Discovery for the RL learning}

Given a set of $N$ users, each user is with a trajectory of $T$
points. Thus in total, we have $NT=N\times T$ tuples $\mtcalD=\left\{ \mtcalD_{n}\mid n=1,\cdots,N\right\} $
for all the $N$ users, where $\mtcalD_{n}=\left\{ \mtcalU_{i}=\left(s_{i},a_{i,}r_{i},s_{i}'\right)\mid i=1,\cdots,T\right\} $
summarizes all the $T$ tuples for the $n$-th user and $\mtcalU_{i}=\left(s_{i},a_{i,}r_{i},s_{i}'\right)$
is the $i$-th tuple in $\mtcalD_{n}$.

\subsection{Pooled-RL and Separate RL (Separ-RL) \label{sub:PooledRL-and-SeparRL}}

The first type of RL methods (i.e. Pooled-RL) assumes that all the
$N$ users are completely homogenous (i.e. same MDPs); they share
all information and run an identical RL for all users\ \cite{PengLiao_2015_Proposal_offPolicyRL}.
In this setting, the critic updating, with an aim of seeking for solutions
to satisfy the Linear Bellman equation\ \cite{Grondman_2012_IEEEts_surveyOfActorCriticRL,Geist_2013_TNNLS_RL_valueFunctionApproximation},
is{\small{}
\begin{equation}
\mtbfw=f\left(\mtbfw\right)=\arg\min_{\mtbfh}\frac{1}{\left|\mtcalD\right|}\sum_{\mtcalU_{i}\in\mtcalD}\left\Vert \mtbfx\left(s_{i},a_{i}\right)^{\intercal}\mtbfh-\left(r_{i}+\gamma\mtbfy\left(s_{i}';\theta\right)^{\intercal}\mtbfw\right)\right\Vert _{2}^{2}+\zeta_{c}\left\Vert \mtbfh\right\Vert _{2}^{2},\label{eq:TD_obj_pooledRL}
\end{equation}
}where $\mtbfw\!=\!f\left(\mtbfw\right)$ is a fixed point problem;
$\mtbfx_{i}\!=\!\mtbfx\left(s_{i},a_{i}\right)^{\intercal}$ is the
value feature at time point $i$; $\mtbfy_{i}=\mtbfy\left(s_{i}';\theta\right)=\sum_{a\in\mtcalA}\mathbf{x}\left(s_{i}',a\right)\pi_{\theta}\left(a\mid s_{i}'\right)$
is the feature at the next time point; $\zeta_{c}$ is a tuning parameter.
The least-square temporal difference for Q-value (LSTD$Q$)\ \cite{Michail_2003_JMLR_LSPI_LSTDQ,AndrewNg_2009_ICML_RLsparity}
provides a closed-form solver for\ \eqref{eq:TD_obj_pooledRL} as
follows {\small{}
\begin{equation}
\mhbfw=\left(\zeta_{c}\mathbf{I}+\frac{1}{\left|\mtcalD\right|}\sum_{\mtcalU_{i}\in\mtcalD}\mtbfx_{i}\left(\mtbfx_{i}-\gamma\mtbfy_{i}\right)^{\intercal}\right)^{-1}\left(\frac{1}{\left|\mtcalD\right|}\sum_{\mtcalU_{i}\in\mtcalD}\mtbfx_{i}r_{i}\right).\label{eq:LSTD_solver_pooledRL}
\end{equation}
}{\small \par}

Since $d_{\text{ref}}\left(s\right)$ is generally unavailable, the
$T$-trial objective for\ \eqref{eq:obj_actorUpdating} is 
\begin{equation}
\hat{\theta}=\arg\max_{\theta}\frac{1}{\left|\mtcalD\right|}\sum_{\mtcalU_{i}\in\mtcalD}\sum_{a\in\mtcalA}Q\left(s_{i},a;\mathbf{\mhbfw}\right)\pi_{\theta}\left(a|s_{i}\right)-\frac{\zeta_{a}}{2}\left\Vert \theta\right\Vert _{2}^{2},\label{eq:Avg_obj_pooledRL}
\end{equation}
where $Q\left(s_{i},a;\mathbf{\mhbfw}\right)=\mathbf{x}\left(s_{i},a\right)^{\intercal}\widehat{\mathbf{w}}$
is the newly estmated Q-value via the critic updating\ \eqref{eq:LSTD_solver_pooledRL}
and $\zeta_{a}$ is the tuning parameter to prevent overfitting\footnote{In case of large feature spaces, one can recursively update $\mhbfw$
via\ \eqref{eq:LSTD_solver_pooledRL} and $\widehat{\theta}$ in\ \eqref{eq:Avg_obj_pooledRL}
to reduce the computational cost. }.

The Pooled-RL works well when all the $N$ users are very similar.
However, there are great behavior discrepancies among users in the
mHealth study because they have different ages, races, incomes, religions,
education levels etc. Such case makes the current Pooled-RL too simple
to simultaneously fit all the $N$ different users' behaviors. It
easily results in lots of biases in the learned value and policy.
The state-of-the-art deep learning methods can be a great idea to
deal with this problem\ \cite{xinliang_2017_CVPR_WSISA,yaoyao_2017_MICCAI,zhengxu_2017_ACMBCB}.

The second type of RL methods (Separ-RL), such as Lei's online contextual
bandit for mHealth\ \cite{huitian_2014_NIPS_ActCriticBandit4JITAI,huitian_2016_PhdThesis_actCriticAlgorithm},
assume that all users are completely heterogeneous. They share no
information and run a separate online RL for each user. The objective
functions are very similar with\ \eqref{eq:TD_obj_pooledRL}, \eqref{eq:LSTD_solver_pooledRL},
\eqref{eq:Avg_obj_pooledRL}.  This method should be great when the
data for each user is very large in size. In this case, the learned
policy is expected be successfully adapted to the user's status. However,
it generally costs a lot (time and other resources) to collect enough
data for the Separ-RL learning. Taking the HeartSteps for example,
it takes 42 days to do the trial collecting 210 tuples per user. What
is worse, there are missing and noises in the data, which will surely
reduce the effective sample size. The problem of small sample size
will easily lead to some unstable policies that contain lots of variances.

\subsection{Group driven RL learning (Gr-RL)}

We observe that users in mHealth are generally similar with some (but
not all) users in the sense that they may have some similar features,
such as age, gender, race, religion, education level, income and other
socioeconomic status\ \cite{Tianxi_2016_CORR_PredictModels4NetworkLinkedData}.
To this end, we propose a group based RL for mHealth to understand
how to share information across similar users to improve the performance.
Specifically, the users are assumed to be grouped together and likely
to share information with others in the same group. The main idea
is to divide the $N$ users into $K$ groups, and learn a separate
RL model for each group. The samples of users in a group are pooled
together, which not only ensures the simplicity of the data for each
RL learning compared with that of the Pooled-RL, but also greatly
enriches the samples for the RL learning compared with that of the
Separ-RL, with an average increase of $\left(N/K-1\right)\times100\%$
on sample size (cf. Section\ \ref{sub:PooledRL-and-SeparRL}). 

To cluster the $N$ users, we employ one of the most benchmark clustering
method, i.e., K-means. The behavior information (i.e. states and rewards)
in the trajectory is processed as the feature. Specifically, the $T$
tuples of a user are stacked together $\mtbfz_{n}=\left[s_{1},r_{1},\cdots,s_{T},r_{T}\right]^{\intercal}$.
With this new feature, we have the objective for clustering as $J=\sum_{n=1}^{N}\sum_{k=1}^{K}r_{nk}\left\Vert \mtbfz_{n}-\bm{\mu}_{k}\right\Vert ^{2}$,
where $\bm{\mu}_{k}$ is the $k$-th cluster center and $r_{nk}\in\left\{ 0,1\right\} $
is the binary indicator variable that describes which of the $K$
clusters the data $\mtbfz_{n}$ belongs to. After the clustering step,
we have the group information $\left\{ \mtcalG_{k}\mid k=1,\cdots,K\right\} $,
each of which includes a set of similar users. With the clustering
results, we have the new objective for the critic updating as $\mtbfw_{k}=f\left(\mtbfw_{k}\right)=\mtbfh_{k}^{*}$,
where $\mtbfh_{k}^{*}$ is estimated from 
\begin{equation}
\min_{\mtbfh_{k}}\frac{1}{\left|\mtcalG_{k}\right|}\sum_{\mtcalU_{i}\in\mtcalG_{k}}\left\Vert \mtbfx{}_{i}^{\intercal}\mtbfh_{k}-\left(r_{i}+\gamma\mtbfy_{i}^{\intercal}\mtbfw_{k}\right)\right\Vert _{2}^{2}+\zeta_{c}\left\Vert \mtbfh_{k}\right\Vert _{2}^{2},\quad\mtfor\ k\in\left\{ 1,\cdots,K\right\} \label{eq:TD_obj_CDRL}
\end{equation}
which could be solved via the LSTD$Q$. The objective for the actor
updating is
\begin{equation}
\max_{\theta_{k}}\frac{1}{\left|\mtcalG_{k}\right|}\sum_{\mtcalU_{i}\in\mtcalG_{k}}\sum_{a\in\mtcalA}Q\left(s_{i},a;\mhbfw_{k}\right)\pi_{\theta_{k}}\left(a|s_{i}\right)-\frac{\zeta_{a}}{2}\left\Vert \theta_{k}\right\Vert _{2}^{2},\quad\mtfor\ k\in\left\{ 1,\cdots,K\right\} .\label{eq:eq:Avg_obj_CDRL}
\end{equation}
By properly setting the value of $K$, we could balance the conflicting
goal of reducing the discrepancy between connected users while increasing
the number of samples for each RL learning: (a) a small $K$ is suited
for the case where $T$ is small and the users are generally similar;
(b) while a large $K$ is adapted to the case where $T$ is large
and users are generally different from others. Besides, we find that
the proposed method is a generalization of the conventional Pooled-RL
and Separ-RL: (a) when $K=1$, the proposed method is equivalent to
the Pooled-RL; (b) when $K=N$, our method is equivalent to the Separ-RL.

\section{Experiments}

There are three RL methods for comparison: (a) the Pooled-RL that
pools the data across all users and learn an identical policy\ \cite{PengLiao_2015_Proposal_offPolicyRL,SusanMurphy_2016_CORR_BatchOffPolicyAvgRwd}
for all users; (b) the Separ-RL, which learns a separate RL policy
for each user by only using his or her data\ \cite{huitian_2014_NIPS_ActCriticBandit4JITAI};
(c) The group driven RL (Gr-RL) is the proposed method. 

The HeartSteps\footnote{i.e., a dataset from the mobile health study, called HeartSteps\ \cite{Walter_2015_Significance_RandomTrialForFitbitGeneration},
to approximate the generative model.} is used for the evaluation. It is a 42-day mHealth intervention that
aims to increase the users' steps they take everyday by providing
some positive suggestions, such as going for a walk after long sitting\ \cite{Walter_2015_Significance_RandomTrialForFitbitGeneration}.
In our study, there are two choices for a policy $\left\{ 0,1\right\} $:
$a=1$ indicates sending the positive intervention, while $a=0$ means
no intervention\ \cite{SusanMurphy_2016_CORR_BatchOffPolicyAvgRwd}.
Specifically, the parameterized stochastic policy is assumed to be
in the form $\pi_{\theta}\left(a\mid s\right)\!=\!\frac{\exp\left[-\theta^{\intercal}\phi\left(s,a\right)\right]}{\sum_{a'}\exp\left[-\theta^{\intercal}\phi\left(s,a\right)\right]}$,
where $\theta\in\mtbbR^{q}$ is the unkown variance and $\phi\left(\cdot,\cdot\right)$
is the feature processing method for the parameterized policy , i.e.,
$\phi\left(s,a\right)=\left[as^{\intercal},a\right]^{\intercal}\in\mtbbR^{m}$.
\begin{table}
\begin{centering}
\caption{{\small{}The average reward of three RL methods when the discount
factor $\gamma$ changes from $0$ to $0.95$: (a) Pooled-RL, (b)
Separ-RL, (c) Gr-RL$_{K=3}$ and Gr-RL$_{K=7}$. A larger value is
better. The }\textbf{\small{}bold value}{\small{} is the best and
the }\emph{\small{}blue italic value}{\small{} is the 2nd best. \label{tab:AverageRwd_PooledRL_SeparRL_CD-RL}}}

\par\end{centering}

\begin{centering}
\begin{tabular}{|c|c|c|c|c|}
\hline 
\multirow{2}{*}{$\gamma$ } &
\multicolumn{4}{c|}{Average reward ($T=42$) }\tabularnewline
\cline{2-5} 
 & Pooled-RL &
Separ-RL &
Gr-RL$_{K=3}$ &
Gr-RL$_{K=7}$\tabularnewline
\hline 
$0$ &
1268.6$\pm$68.2 &
1255.3$\pm$62.3 &
\textit{\textcolor{blue}{1279.0$\pm$66.6}} &
\textbf{1289.5$\pm$64.5}\tabularnewline
$0.2$ &
1268.1$\pm$68.3 &
1287.6$\pm$76.8 &
\textit{\textcolor{blue}{1318.3$\pm$62.5}} &
\textbf{1337.3$\pm$56.7}\tabularnewline
$0.4$ &
1267.6$\pm$68.4 &
1347.0$\pm$54.1 &
\textit{\textcolor{blue}{1368.8$\pm$57.6}} &
\textbf{1389.7$\pm$50.7}\tabularnewline
$0.6$ &
1267.3$\pm$68.5 &
1357.6$\pm$57.9 &
\textit{\textcolor{blue}{1441.3$\pm$48.2}} &
\textbf{1446.3$\pm$46.7}\tabularnewline
$0.8$ &
1266.8$\pm$68.7 &
1369.4$\pm$51.6 &
\textbf{1513.9$\pm$38.8} &
\textit{\textcolor{blue}{1484.0$\pm$44.5}}\tabularnewline
$0.95$ &
\multicolumn{1}{c|}{1266.3$\pm$68.7} &
1348.9$\pm$53.4 &
\textbf{1538.6$\pm$34.3} &
\textit{\textcolor{blue}{1500.6$\pm$42.8}}\tabularnewline
\hline 
Avg. &
1267.4 &
1327.6 &
\textbf{1410.0} &
\textit{\textcolor{blue}{1407.9}}\tabularnewline
\hline 
\hline 
$\gamma$  &
\multicolumn{4}{c|}{Average reward ($T=100$) }\tabularnewline
\hline 
$0$ &
1284.4$\pm$64.1 &
1271.1$\pm$70.7 &
\textit{\textcolor{blue}{1293.5$\pm$62.1}} &
\textbf{1294.9$\pm$63.7}\tabularnewline
$0.2$ &
1285.8$\pm$63.9 &
1301.2$\pm$65.6 &
\textit{\textcolor{blue}{1329.6$\pm$58.5}} &
\textbf{1332.9$\pm$58.7}\tabularnewline
$0.4$ &
1287.1$\pm$63.8 &
1370.1$\pm$49.1 &
\textit{\textcolor{blue}{1385.5$\pm$52.1}} &
\textbf{1393.0$\pm$49.2}\tabularnewline
$0.6$ &
1288.5$\pm$63.6 &
1409.3$\pm$42.2 &
\textit{\textcolor{blue}{1452.9$\pm$44.3}} &
\textbf{1459.6$\pm$40.9} \tabularnewline
$0.8$ &
1289.9$\pm$63.4 &
1435.0$\pm$37.6 &
\textbf{1519.0$\pm$39.5} &
\textit{\textcolor{blue}{1518.0$\pm$38.5}}\tabularnewline
$0.95$ &
1291.2$\pm$63.2 &
1441.9$\pm$35.9 &
\textbf{1547.2$\pm$37.2} &
\textit{\textcolor{blue}{1540.6$\pm$38.1}}\tabularnewline
\hline 
Avg. &
1287.8 &
1371.4 &
\textit{\textcolor{blue}{1421.3}} &
\textbf{1423.2}\tabularnewline
\hline 
\end{tabular}
\par\end{centering}

\emph{\footnotesize{}The value of $\gamma$ specifies different RL
methods: (a) $\gamma=0$ means the contextual bandit\ \cite{huitian_2014_NIPS_ActCriticBandit4JITAI},
(b) $0<\gamma<1$ indicates the discounted reward RL. \vspace{-0.45cm}}{\footnotesize \par}
\end{table}

\subsection{Experiments Settings}

A trajectory of $T$ tuples $\mathcal{D}_{T}=\left\{ \left(s_{i},a_{i},r_{i}\right)\right\} _{i=1}^{T}$
are collected from each user via the micro-randomized trial\ \cite{SusanMurphy_2016_CORR_BatchOffPolicyAvgRwd,huitian_2014_NIPS_ActCriticBandit4JITAI}.
The initial state is sampled from the Gaussian distribution $S{}_{0}\sim\mtcalN_{p}\left\lbrace 0,\Sigma\right\rbrace $,
where $\Sigma$ is the $p\times p$ covariance matrix with pre-defined
elements. The policy of selecting action $a_{t}=1$ is drawn from
the random policy with a probability of $0.5$ to provide interventions,
i.e. $\mu\left(1\mid s_{t}\right)=0.5$ for all states $s_{t}$. For
$t\geq1$, the state and immediate reward are generated as follows
\begin{align}
S_{t,1} & =\beta_{1}S_{t-1,1}+\xi_{t,1},\nonumber \\
S_{t,2} & =\beta_{2}S_{t-1,2}+\beta_{3}A_{t-1}+\xi_{t,2},\label{eq:Dat=0000231_stateTrans_cmp3}\\
S_{t,3} & =\beta_{4}S_{t-1,3}+\beta_{5}S_{t-1,3}A_{t-1}+\beta_{6}A_{t-1}+\xi_{t,3},\nonumber \\
S_{t,j} & =\beta_{7}S_{t-1,j}+\xi_{t,j},\qquad\mtfor\ j=4,\ldots,p\nonumber \\
R_{t} & =\beta_{14}\times\left[\beta_{8}+A_{t}\times(\beta_{9}+\beta_{10}S_{t,1}+\beta_{11}S_{t,2})+\beta_{12}S_{t,1}-\beta_{13}S_{t,3}+\varrho_{t}\right],\label{eq:Dat=0000231_ImmediateRwd_cmp3}
\end{align}
where $\bm{\beta}=\left\{ \beta_{i}\right\} _{i=1}^{14}$ are the
main parameters for the MDP; $\left\{ \xi_{t,i}\right\} _{i=1}^{p}\sim\mtcalN\left(0,\sigma_{s}^{2}\right)$
is the noise in the state\ \eqref{eq:Dat=0000231_ImmediateRwd_cmp3}
and $\varrho_{t}\sim\mtcalN\left(0,\sigma_{r}^{2}\right)$ is the
noise in the reward model\ \eqref{eq:Dat=0000231_ImmediateRwd_cmp3}.
To simulate $N$ users that are similar but not identical, we need
$N$ different $\bm{\beta}$s, each of which is similar with a set
of others. Formally, there are two steps to obtain $\bm{\beta}$ for
the $i$-th user: (a) select the $m$-th basic $\bm{\beta}$, i.e.
$\bm{\beta}_{m}^{\text{basic}}$, which determines which group the
$i$-th user belong to; (b) add the noise $\bm{\beta}_{i}=\bm{\beta}_{m}^{\text{basic}}+\bm{\delta}_{i},\ \text{for}\ i\in\left\{ 1,2,\cdots,N_{m}\right\} $
to make each user different from others, where $N_{m}$ indicates
the number of users in the $m$-th group, $\bm{\delta}_{i}\sim\mtcalN\left(0,\sigma_{b}\mtbfI_{14}\right)$
is the noise and $\mtbfI_{14}\in\mtbbR^{14\times14}$ is an identity
matrix. The value of $\sigma_{b}$ specifies how different the users
are. Specially in our experiment, we set $M=5$ groups (each group
has $N_{m}=10$ people, leading to $N=50$ users involved in the experiment).
The basic $\bm{\beta}$s for the $M$ groups are set as follows {\scriptsize{}
\begin{align*}
\bm{\beta}_{1}^{\text{basic}} & =\left[0.40,0.25,0.35,0.65,0.10,0.50,0.22,2.00,0.15,0.20,0.32,0.10,0.45,800\right]\\
\bm{\beta}_{2}^{\text{basic}} & =\left[0.45,0.35,0.40,0.70,0.15,0.55,0.30,2.20,0.25,0.25,0.40,0.12,0.55,700\right]\\
\bm{\beta}_{3}^{\text{basic}} & =\left[0.35,0.30,0.30,0.60,0.05,0.65,0.28,2.60,0.35,0.45,0.45,0.15,0.50,650\right]\\
\bm{\beta}_{4}^{\text{basic}} & =\left[0.55,0.40,0.25,0.55,0.08,0.70,0.26,3.10,0.25,0.35,0.30,0.17,0.60,500\right]\\
\bm{\beta}_{5}^{\text{basic}} & =\left[0.20,0.50,0.20,0.62,0.06,0.52,0.27,3.00,0.15,0.15,0.50,0.16,0.70,450\right],
\end{align*}
}Besides, the noises are set $\sigma_{s}=\sigma_{r}=1$ and $\sigma_{\beta}=0.01$.
Other variances are $p=3$, $q=4$, $\zeta_{a}=\zeta_{c}=0.01$. The
feature procesing for the value estimation is $\mtbfx\left(s,a\right)=\left[1,s^{\intercal},a,s^{\intercal}a\right]^{\intercal}\in\mtbbR^{2p+2}$
for all the compared methods.

\subsection{Evaluation Metric and Results \label{sub:EvaluationMetric-and-Results}}

In the experiments, the expectation of long run average reward (ElrAR)
$\mathbb{E}\left[\eta^{\pi_{\hat{\theta}}}\right]$ is proposed to
evaluate the quality of a learned policy $\pi_{\hat{\theta}}$\ \cite{PengLiao_2015_Proposal_offPolicyRL,SusanMurphy_2016_CORR_BatchOffPolicyAvgRwd}.
Intuitively in the HeartSteps application, ElrAR measures the average
step a user could take each day when he or she is provided by the
intervention via the learned policy $\pi_{\hat{\theta}}$. Specifically,
there are two steps to achieve the ElrAR\ \cite{SusanMurphy_2016_CORR_BatchOffPolicyAvgRwd}:
(a) get the $\eta^{\pi_{\hat{\theta}}}$ for each user by averaging
the rewards over the last $4,000$ elements in the long run trajectory
with a total number of $5,000$ tuples; (a) ElrAR $\mathbb{E}\left[\eta^{\pi_{\hat{\theta}}}\right]$
is achieved by averaging over the $\eta^{\pi_{\hat{\theta}}}$'s of
all users. 
\begin{figure}[t]
\begin{centering}
\includegraphics[width=0.95\linewidth]{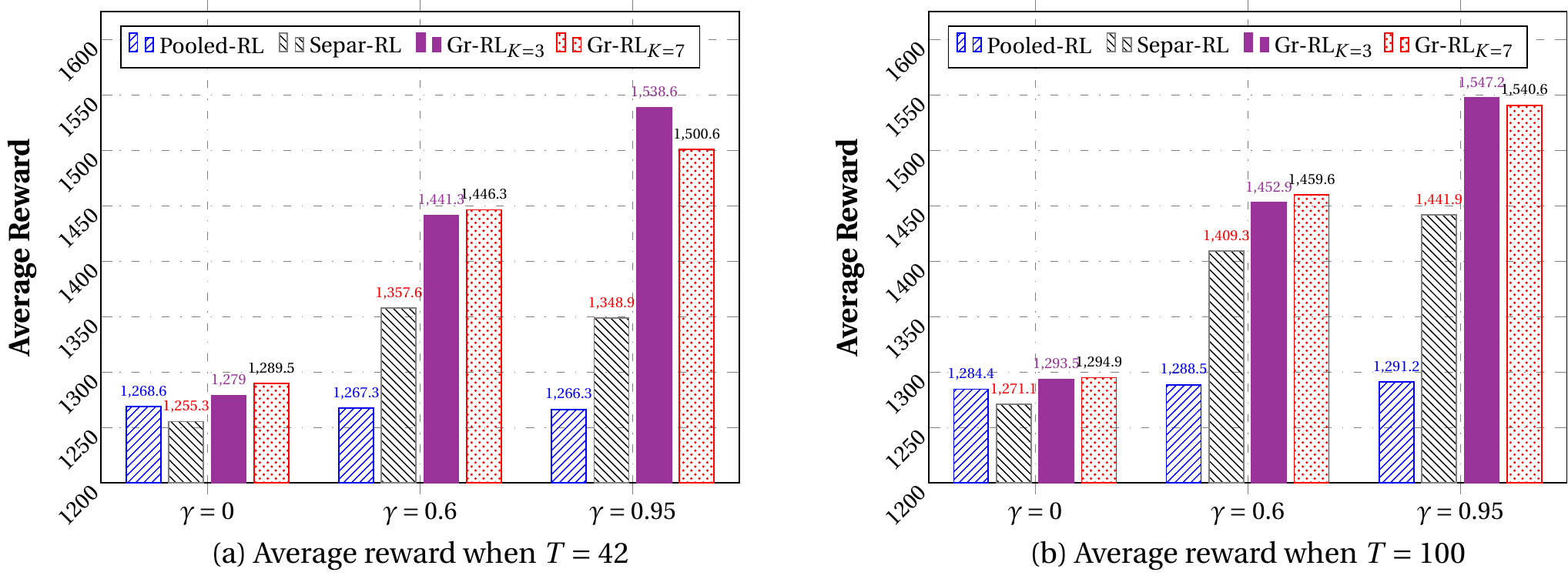}\caption{{\small{}Average reward of 3 RL methods: a) Pooled-RL, (b) Separ-RL,
(c) Gr-RL$_{K=3}$ and Gr-RL$_{K=7}$. The left sub-figure shows the
results when the trajectory is short, i.e. $T=42$; the right one
shows the results when $T=100$. A larger value is better.\label{fig:bar_3_gammas}}}

\par\end{centering}

\vspace{-0.5cm}
\end{figure}

The experiment results of three RL methods are summarized in Table\ \ref{tab:AverageRwd_PooledRL_SeparRL_CD-RL}
and Fig.\ \ref{fig:bar_3_gammas}, that is: (a) Pooled-RL, (b) Separ-RL,
(c) Gr-RL$_{K=3}$ and Gr-RL$_{K=7}$. $K=3,7$ is the number of cluster
centers in our algorithm, which is set different from the true number
of groups $M=5$. Such setting is to show that Gr-RL does not require
the true value of $M$. There are two sub-tables in Table\ \ref{tab:AverageRwd_PooledRL_SeparRL_CD-RL}.
The top sub-table summarizes the experiment results of three RL methods
under six $\gamma$ settings (i.e. the discount reward) when the trajectory
is short, i.e. $T=42$. While the bottom one displays the results
when the trajectory is long, i.e. $T=100$. Each row shows the results
under one discount factor, $\gamma=0,\cdots,0.95$; the last row shows
the average perforance over all the six $\gamma$ settings. 

As we shall see, Gr-RL$_{K=3}$ and Gr-RL$_{K=7}$ generally perform
similarly and are always among the best. Such results demonstrate
that our method doesn't require the true value of groups and is robust
to the value of $K$\ \cite{fyzhu_2014_JSTSP_RRLbS,yingWang_2015_TIP_RobustUnmixing,guangliangCheng_2016_JStars_robustHyperClassification,fyzhu_2014_AAAI_ARSS,xiaoping_2017_ICASSP}.
In average, the proposed method improves the ElrAR by $82.4$ and
$80.3$ steps when $T=42$ as well as $49.8$ and $51.7$ steps when
$T=100$, compared with the best result of the state-of-the-art methods,
i.e. Separ-RL. There are two interesting observations: (1) the improvement
of our method decreases as the trajectory length $T$ increases; (2)
when the trajectory is short, i.e. $T=42$, it is better to set small
$K$s, which emphasizes the enriching of dataset; while the trajectory
is long, i.e. $T=100$, it is better to set large $K$s to simplify
the data for each RL learning.

\section{Conclusions and Discussion}

In this paper, we propose a novel group driven RL method for the mHealth.
Compared with the state-of-the-art RL methods for mHealth, it is based
on a more practical assumption that admits the discrepancies between
users and assumes that a user should be similar with some (but not
all) users. The proposed method is able to balance the conflicting
goal of reducing the discrepancy between pooled users while increasing
the number of samples for each RL learning. Extensive experiment results
verify that our method gains obvious advantages over the state-of-the-art
RL methods in the mHealth.  

{\footnotesize{}\bibliographystyle{3_home_fyzhu_DATA_Dropbox_self_Folder_myWorksOn____GroupDrivenRL_4_mHealth_batchLearning_ieee}
\phantomsection\addcontentsline{toc}{section}{\refname}\bibliography{1_home_fyzhu_link2dropbox_self_Folder_myWorksOnDropboxs_bibFiles_referenceBib,2_home_fyzhu_link2dropbox_self_Folder_myWorksOnDropboxs_bibFiles_referenceBib2}
}{\footnotesize \par}
\end{document}